# SOIL CLASSIFICATION USING GATREE

P.Bhargavi[1], Dr.S.Jyothi[2]

[1] Department of CSE, Madanapalli Institue of Tecnology and Science, Madanapalli,
pbhargavi18@yahoo.co.in

[2] Department of Computer Science, Sri Padmavathi Mahila VisvaVidyalayam,
Tirupati(Womens University), jyothi.spmvv@gmail.com

*ABSTRACT*

*This paper details the application of a genetic programming framework for classification of decision tree of Soil data to classify soil texture. The database contains measurements of soil profile data. We have applied GATree for generating classification decision tree. GATree is a decision tree builder that is based on Genetic Algorithms (GAs). The idea behind it is rather simple but powerful. Instead of using statistic metrics that are biased towards specific trees we use a more flexible, global metric of tree quality that try to optimize accuracy and size. GATree offers some unique features not to be found in any other tree inducers while at the same time it can produce better results for many difficult problems. Experimental results are presented which illustrate the performance of generating best decision tree for classifying soil texture for soil data set.*

*KEY WORDS:*

*Data Mining, Soil Profile, Soil Database, Classification, GATree.*

## 1. INTRODUCTION

Fundamental analysis involves the analysis of economic data, industry conditions, company fundamentals, and corporate financial statements [5]. Data mining consists of the extraction of interesting novel knowledge from real-world databases [1]. Near boundless effort is expended in analyzing time series consisting of market and company metrics to predict future outcomes in order to achieve above average returns. This paper details an application of genetic programming to the problem of obtaining interesting knowledge from the soil dataset[7]. The database consisting measurements of soil profile data from various locations of Rayalaseema Region. Here we propose a genetic programming framework for induction of classification from databases. The framework outlines a method for classification of soil texture for soil data set using Genetic Algorithm. Experimental results are presented which illustrate the performance of generating the best decision tree for classifying soil texture for soil data set.

             

## 2. SOIL CLASSIFICATION

Of soil characteristics, soil Classification is the most important one. It influences many other properties of great significance to land use and management. The Soil texture is an important property for agriculture soil classification. It influences fertility, drainage, water holding capacity, aeration, tillage, and strength of soils.

A set of soil properties are diagnostic for differentiation of pedons. The differentiating characters are the soil properties that can be observed in the field or measured in the laboratory or can be inferred in the field. Some diagnostic soil horizons, both surface and sub- surfaces, soil moisture regimes, soil temperature regimes and physical, physio-chemical and chemical properties of soils determined were used as criteria for classifying soils. The soils of various regions are classified into different orders, sub-orders, great groups, sub-groups, families and finally into series as per USDA Soil Taxonomy[14]. The texture of the surface varied from sand to silty clay loam where as in sub-surface horizons it varied from sand to clay[7].

The solid phase of soil can be divided into mineral matter and organic matter. The mineral particles can be futher subdivided into classes based on size. The classification of soil particles according to size are Sand, Silt, Clay. The proposition of Sand, Silt, Clay present in soil determines its textue.

### 2.1 SOIL DATA

In this paper Soil data consists of attributes like (i.e., Depth, Sand, Silt, Clay, Sandbysilt, Sandbyclay, Sandbysiltclay, TextureClass). The texture of the Soil data is varied from sand to silty clay loam where as in sub-surface horizons it varied from sand to clay[2]. Table 1 shows the different soil survey symbols.

**Table 1: Soil Survey Symbols**

| | |
|---|---|
| **S** | **Sand** |
| **Sicl** | **Silty Clay Loam** |
| **Sic** | **Silty Clay** |
| **C** | **Clay** |
| **Sl** | **Sandy loam** |
| **Cl** | **Clay loam** |
| **Sil** | **Silty Loam** |
| **L** | **Loam** |
| **Ls** | **Loamy sand** |
| **Scl** | **Sand Clay Loam** |
| **Sc** | **Sand Clay** |

## 3. GENETIC ALGORITHM

Genetic Algorithm is the method for selecting the most suitable answer by using feasibility and Natural Selection of Charles Darwin [9]. Genetic Algorithm (GA) has been developed during the 60th decade and has become quite popular from being distributed by John Holland who published the book called, "Adaptation in Natural and Artificial Systems" for the first time in 1975. The process of GA was copied from natural selection that could be explained as the replacement of interesting problems by string of numbers or in biology as chromosomes. Each chromosome contained gene which was replaced by Decision Variable. In the first place, gene would be randomly selected to choose the population size. Later, each chromosome had been evaluated for Objective Function for fitness which represented the value of suitability of chromosomes before entering the process of GA through selection to find origin of species.

In this paper soil classification is performed using GATree [6], which is a decision tree builder that is based on Genetic Algorithms (GAs). The idea behind it is rather simple but powerful. Instead of using statistic metrics that are biased towards specific trees we use a more flexible, global metric of tree quality that try to optimize accuracy and size. GATree offers some unique features not to be found in any other tree inducers while at the same time it can produce better results for many difficult problems.

The main screen of the GA tree is shown in figure 1.

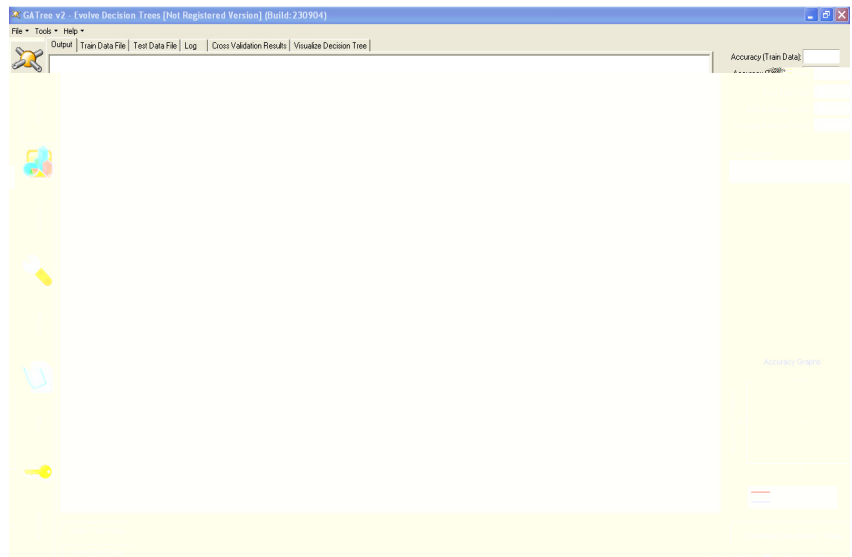

**Figure 1: Main Screen of GATree tool.**

The main screen of the program (Figure 1) allows us to select an active training dataset and evolve the decision tree. In the main program's window we can watch the best decision tree as it evolves through time. The right panel includes information about the current status of the evolution process.

GATree uses ARFF as its standard source format. An ARFF file is a simple text file that describes the problem instances and its attributes.

By pressing the Visualize-Decision-Tree button we can visualize and traverse the final decision tree

The statistics tab on the main screen provides several graphs of the evolution process. Those graphs allow us to follow the evolution process in real time and discover potential problems and trends. As an example, when the Average Fitness of the population tend to be equal to the Fitness of the best Genome then there is little room for further improvements. A solution here could be to try with more generations or bigger population size.

The settings tab on the main screen allows us to control every aspect of the evolution process. There are two types of settings; Basic settings and advanced settings depending on their usefulness and complexity. Below you can find an explanation of the offered options for Soil dataset.

**Selection:**

The selection operation for Soil data is performed by selecting the best chromosome in a given population of 100. The decision tree is evolved at each generation with the Genome Score of 0.34. At each iteration the decision tree is reproduced by combing the best genomes of parents to produce the best genomes of child. the best individuals in a population is determined with a fitness value of 0.28, which is used in the selection process to choose parents. The process continues until all individuals are selected from a population with 100 generations. They effectively control the total time for evolution of decision tree.

**Crossover:**

With the given attributes of Soil data like (i.e., Depth, Sand, Silt, Clay, Sandbysilt, Sandbyclay, Sandbysiltclay) as parents, the crossover is generated with new individuals of textureclass like ((i.e. s, sicl, sic, c, sl, cl, sil, l, ls, scl), as child nodes. The probability that a random subtree is replaced with another subtree is 0.99.

**Mutation:**

As the tree evolves from the root node i.e parent to the leaf node i.e children, randomly it changes the characters in the children. For Soil data the decision tree is evolved with the input attribute if sand <= 54 and outputs two nodes i.e if the condition is true it outputs if clay <= 8 and if the condition is false it outpyts if sand <= 79, which again generates the child nodes.Until it outputs the texture of soil with a very small mutation probability 0.01, which refers to the probability for a node to be randomly altered to include a new value.

## 4. EXPERIMENTAL RESULTS

The GATree was applied on soil data set and the results obtained are shown below:

A decision tree with a population of 100 was generated with an average size of 49. The decision tree was evolved with an accuracy of 0.59 for each generation. The number of bad trees that will be replaced with new ones between generations is 0.25. The probability of replacing a random subtree with another subtree is 0.99 and the probability for a node to be randomly altered to include a new value is 0.01, with these probabilities the speed and accuracy of evolution of decision tree is increased to an average accuracy of 0.45.

With the cross validation the accuracy is estimated to generate a more accurate decision tree. The performance of decision tree is more accurate to classify the soil data set. The accuracy graph is shown in figure 2.

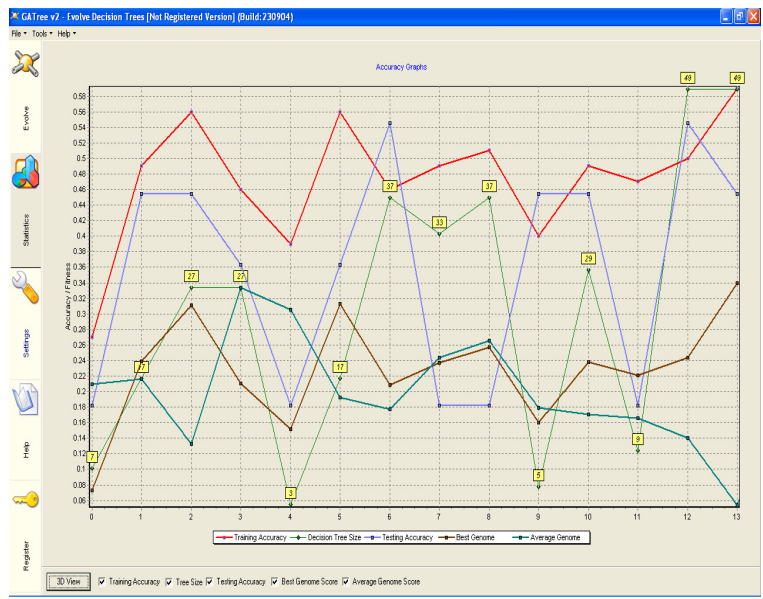

**Figure 2: Accuracy Graph for Soil Data Set .**

In the above graph blue line indicates testing accuracy, red line indicates training accuracy, light green line indicates decision tree size, brown indicates best genome, dark green indicates average genome. The testing accuracy line is close to training accuracy line so our model fits not only for training data but also for testing data. The decision tree of height 13 was generated, but the pruned decision tree is shown figure 3.

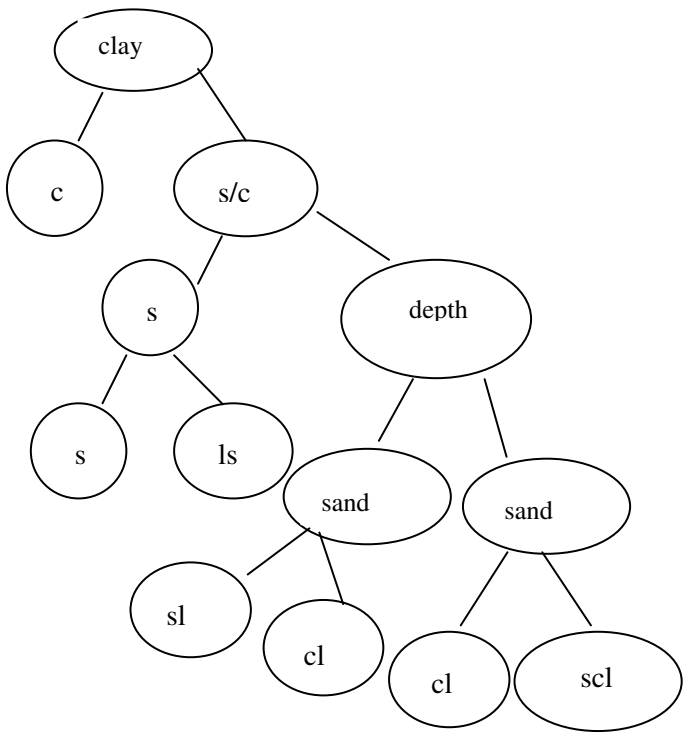

**Figure 3: The Pruned Decision Tree**

In the above decision tree the label s/c in the node denotes the ratio of silt by clay, c denotes clay, s dentoes sandy, ls denotes loamy sand, sl denotes sandy loam, cl denotes clay loam, scl denotes sandy clay loam.
The rules are analyzed for the above decision tree as shown below:

>If  clay <=34
>>Then texture =c
>
>Else
>>If s/c <=3
>>>Then texture=s
>>>If s<=10

```
                                Then texture=s
                        Else
                                Texture=ls
                Else
                        If depth<=0.55
                                Then if sand <=45
                                        Then texture=sl
                                Else
                                        Texture=cl
                        Else
                                Then if sand<=82
                                        Then texture=cl
                                Else
                                        Texture=scl
```

## 5. CONCLUSION

Thus this paper discusses on how genetic algorithms can be applied to soil data and how the genetic algorithms are used to generate the best decision tree. We can say that the average accuracy gets improved for bigger decision trees with more number of nodes than a smaller decision tree with less number of nodes. Further we can conclude that the genetic algorithm was applied on smaller soil data set to classify soil texture. It is evident from the above example that the Average Fitness of the population tends to be equal to the Fitness of the best Genome so, there is little room for further improvements. In the future we can try applying genetic algorithms on bigger population size for soil data set and can be tried to check whether we get accurate results to classify soil texture and whether more accurate rules can be generated than data mining classification algorithms. Applying Genetic Algorithm helps one to classify soil texture based on soil properties effectively, which influences fertility, drainage, water holding capacity, aeration, tillage, and bearing strength of soils and also helps in knowing the accuracy of a decision tree along with accurate rules.

## Authors

**Bhargavi Peyakunta** is working as Associate Professor in the Department of Computer Science and Engineering, Madanapalle Institute of Technology and Science, Madanapalle, Andhra Pradesh. **Educational Qualifications:** M.Sc in Computer Science from Sri Krishnadevaraya University, Anantapur and M.Tech degree from Sri Vinayaka missions University, Salem, India. **Teaching & Research Experience:** 12 years of teaching experience & 5 years of research experience. **Current Research Interests:** Data Mining, Fuzzy Systems, Genetic Algorithms and GIS.

**Jyothi Singaraju** is working as Head & Associate Professor in the Department of Computer Science, Sri Padmavathi Mahila Visvavidyalayam(SPMVV), Tirupati. **Educational Qualifications:** M.Sc in Applied Mathematics from S.V.University, Tirupati , M.S in Software Systems from BITS, Pilani, & Ph.D in Theoritical Computer Science from S.V.University, Tirupati. **Teaching & Research Experience:** 16 years teaching experience & 24 years research experience. **Current Research Interests:** Fuzzy Systems, Neural Networks, Data Mining, Data Base Management Systems, Genetic Algorithms, Bioinformatics and GIS.